\documentclass[10pt,twocolumn,letterpaper]{article}

\usepackage{wacv}

\makeatletter
\@namedef{ver@everyshi.sty}{}
\makeatother

\usepackage{times}
\usepackage{graphicx}
\usepackage{multirow}
\usepackage{amsmath}
\usepackage{amssymb}
\usepackage{booktabs}
\usepackage{soul}
\usepackage{xcolor}
\usepackage{colortbl}
\usepackage{enumitem}
\usepackage{flushend}

\usepackage{tikz}
\usetikzlibrary{arrows,shapes,calc,matrix,fit,backgrounds}
\usepackage{pgfplots}
\usepackage{pgfplotstable}
\pgfplotsset{compat=1.9}
\usepgfplotslibrary{fillbetween}

\usepackage{xstring}

\usepgfplotslibrary{external}

\IfBeginWith*{\jobname}{fig/extern/}{\finalcopy}{}

\tikzset{every mark/.append style={solid}}
\pgfplotsset{
	grid=both, width=\columnwidth, try min ticks=5,
	every axis/.append style={font=\scriptsize},
	every axis plot/.append style={thick,mark=none,mark size=1.2,tension=0.18},
	legend cell align=left, legend style={fill opacity=0.8},
}

\pgfplotsset{
	dash/.style={mark=o,dashed,opacity=0.7},
	dott/.style={mark=o,dotted,opacity=0.7},
}

\wacvfinalcopy 

\ifwacvfinal
\usepackage[breaklinks=true,bookmarks=false]{hyperref}
\else
\usepackage[pagebackref=true,breaklinks=true,colorlinks,bookmarks=false]{hyperref}
\fi

\begin{document}

\title{Large-to-small Image Resolution Asymmetry in Deep Metric Learning}

\author{Pavel Suma \hspace{10ex} Giorgos Tolias\\
Visual Recognition Group, Faculty of Electrical Engineering, Czech Technical University in Prague\\
{\tt\small sumapave,toliageo@fel.cvut.cz}
}

\maketitle
\thispagestyle{empty}

\newcommand{\mypartight}[1]{\noindent {\bf #1}}
\newcommand{\myparagraph}[1]{\vspace{3pt}\noindent\textbf{#1}\xspace}

\newcommand{\alert}[1]{{\color{red}{#1}}}
\newcommand{\gt}[1]{{\color{purple}{GT: #1}}}
\newcommand{\gtt}[1]{{\color{purple}{#1}}}
\newcommand{\gtr}[2]{{\color{purple}\st{#1} {#2}}}

\newcommand{\ps}[1]{{\color{blue}{PS: #1}}}
\newcommand{\pst}[1]{{\color{blue}{#1}}}
\newcommand{\psr}[2]{{\color{blue}\st{#1} {#2}}}

\newcommand{\gray}[1]{{\color{gray}{#1}}}

\newcommand{\symcol}{olive}
\newcommand{\asmrescol}{teal}
\newcommand{\asmdrescol}{purple}
\newcommand{\asmnetcol}{orange}
\newcommand{\asmdbothcol}{violet}
\newcommand{\bothmark}{*}
\newcommand{\netmark}{triangle*}
\newcommand{\resmark}{diamond*}
\newcommand{\dresmark}{pentagon*}
\newcommand{\dbothmark}{square*}

\def\roxf{$\mathcal{R}$Oxford\xspace}
\def\rox{$\mathcal{R}$Oxf\xspace}
\def\ro{$\mathcal{R}$O\xspace}
\def\rpar{$\mathcal{R}$Paris\xspace}
\def\rpa{$\mathcal{R}$Par\xspace}
\def\rp{$\mathcal{R}$P\xspace}
\def\rdis{$\mathcal{R}$1M\xspace}

\newcommand\resnet[3]{\ensuremath{\prescript{#2}{}{\mathtt{R}}{#1}_{\scriptscriptstyle #3}}\xspace}

\newcommand{\ours}{Beat-them-all\xspace} 

\newcommand{\stddev}[1]{\scriptsize{$\pm#1$}}

\newcommand{\diffup}[1]{{\color{OliveGreen}{($\uparrow$ #1)}}}
\newcommand{\diffdown}[1]{{\color{BrickRed}{($\downarrow$ #1)}}}

\newcommand{\comment} [1]{{\color{orange} \Comment     #1}} 

\def\nmsp{\hspace{-6pt}}
\def\nssp{\hspace{-3pt}}
\def\nxssp{\hspace{-1pt}}
\def\zsp{\hspace{0pt}}
\def\xssp{\hspace{1pt}}
\def\ssp{\hspace{3pt}}
\def\msp{\hspace{6pt}}
\def\mlsp{\hspace{9pt}}
\def\lsp{\hspace{12pt}}
\def\xlsp{\hspace{20pt}}


\newcommand{\head}[1]{{\smallskip\noindent\bf #1}}
\newcommand{\equ}[1]{(\ref{equ:#1})\xspace}


\newcommand{\nn}[1]{\ensuremath{\text{NN}_{#1}}\xspace}
\def\l1{\ensuremath{\ell_1}\xspace}
\def\l2{\ensuremath{\ell_2}\xspace}


\newcommand{\tran}{^\top}
\newcommand{\mtran}{^{-\top}}
\newcommand{\zcol}{\mathbf{0}}
\newcommand{\zrow}{\zcol\tran}

\newcommand{\ind}{\mathds{1}}
\newcommand{\expect}{\mathbb{E}}
\newcommand{\nat}{\mathbb{N}}
\newcommand{\zahl}{\mathbb{Z}}
\newcommand{\real}{\mathbb{R}}
\newcommand{\proj}{\mathbb{P}}
\newcommand{\prob}{\mathbf{Pr}}

\newcommand{\mif}{\textrm{if }}
\newcommand{\other}{\textrm{otherwise}}
\newcommand{\minimize}{\textrm{minimize }}
\newcommand{\maximize}{\textrm{maximize }}

\newcommand{\id}{\operatorname{id}}
\newcommand{\const}{\operatorname{const}}
\newcommand{\sgn}{\operatorname{sgn}}
\newcommand{\var}{\operatorname{Var}}
\newcommand{\mean}{\operatorname{mean}}
\newcommand{\trace}{\operatorname{tr}}
\newcommand{\diag}{\operatorname{diag}}
\newcommand{\vect}{\operatorname{vec}}
\newcommand{\cov}{\operatorname{cov}}

\newcommand{\softmax}{\operatorname{softmax}}
\newcommand{\clip}{\operatorname{clip}}

\newcommand{\defn}{\mathrel{:=}}
\newcommand{\peq}{\mathrel{+\!=}}
\newcommand{\meq}{\mathrel{-\!=}}

\newcommand{\floor}[1]{\left\lfloor{#1}\right\rfloor}
\newcommand{\ceil}[1]{\left\lceil{#1}\right\rceil}
\newcommand{\inner}[1]{\left\langle{#1}\right\rangle}
\newcommand{\norm}[1]{\left\|{#1}\right\|}
\newcommand{\frob}[1]{\norm{#1}_F}
\newcommand{\card}[1]{\left|{#1}\right|\xspace}
\newcommand{\diff}{\mathrm{d}}
\newcommand{\der}[3][]{\frac{d^{#1}#2}{d#3^{#1}}}
\newcommand{\pder}[3][]{\frac{\partial^{#1}{#2}}{\partial{#3^{#1}}}}
\newcommand{\ipder}[3][]{\partial^{#1}{#2}/\partial{#3^{#1}}}
\newcommand{\dder}[3]{\frac{\partial^2{#1}}{\partial{#2}\partial{#3}}}

\newcommand{\wb}[1]{\overline{#1}}
\newcommand{\wt}[1]{\widetilde{#1}}

\newcommand{\cA}{\mathcal{A}}
\newcommand{\cB}{\mathcal{B}}
\newcommand{\cC}{\mathcal{C}}
\newcommand{\cD}{\mathcal{D}}
\newcommand{\cE}{\mathcal{E}}
\newcommand{\cF}{\mathcal{F}}
\newcommand{\cG}{\mathcal{G}}
\newcommand{\cH}{\mathcal{H}}
\newcommand{\cI}{\mathcal{I}}
\newcommand{\cJ}{\mathcal{J}}
\newcommand{\cK}{\mathcal{K}}
\newcommand{\cL}{\mathcal{L}}
\newcommand{\cM}{\mathcal{M}}
\newcommand{\cN}{\mathcal{N}}
\newcommand{\cO}{\mathcal{O}}
\newcommand{\cP}{\mathcal{P}}
\newcommand{\cQ}{\mathcal{Q}}
\newcommand{\cR}{\mathcal{R}}
\newcommand{\cS}{\mathcal{S}}
\newcommand{\cT}{\mathcal{T}}
\newcommand{\cU}{\mathcal{U}}
\newcommand{\cV}{\mathcal{V}}
\newcommand{\cW}{\mathcal{W}}
\newcommand{\cX}{\mathcal{X}}
\newcommand{\cY}{\mathcal{Y}}
\newcommand{\cZ}{\mathcal{Z}}

\newcommand{\vA}{\mathbf{A}}
\newcommand{\vB}{\mathbf{B}}
\newcommand{\vC}{\mathbf{C}}
\newcommand{\vD}{\mathbf{D}}
\newcommand{\vE}{\mathbf{E}}
\newcommand{\vF}{\mathbf{F}}
\newcommand{\vG}{\mathbf{G}}
\newcommand{\vH}{\mathbf{H}}
\newcommand{\vI}{\mathbf{I}}
\newcommand{\vJ}{\mathbf{J}}
\newcommand{\vK}{\mathbf{K}}
\newcommand{\vL}{\mathbf{L}}
\newcommand{\vM}{\mathbf{M}}
\newcommand{\vN}{\mathbf{N}}
\newcommand{\vO}{\mathbf{O}}
\newcommand{\vP}{\mathbf{P}}
\newcommand{\vQ}{\mathbf{Q}}
\newcommand{\vR}{\mathbf{R}}
\newcommand{\vS}{\mathbf{S}}
\newcommand{\vT}{\mathbf{T}}
\newcommand{\vU}{\mathbf{U}}
\newcommand{\vV}{\mathbf{V}}
\newcommand{\vW}{\mathbf{W}}
\newcommand{\vX}{\mathbf{X}}
\newcommand{\vY}{\mathbf{Y}}
\newcommand{\vZ}{\mathbf{Z}}

\newcommand{\va}{\mathbf{a}}
\newcommand{\vb}{\mathbf{b}}
\newcommand{\vc}{\mathbf{c}}
\newcommand{\vd}{\mathbf{d}}
\newcommand{\ve}{\mathbf{e}}
\newcommand{\vf}{\mathbf{f}}
\newcommand{\vg}{\mathbf{g}}
\newcommand{\vh}{\mathbf{h}}
\newcommand{\vi}{\mathbf{i}}
\newcommand{\vj}{\mathbf{j}}
\newcommand{\vk}{\mathbf{k}}
\newcommand{\vl}{\mathbf{l}}
\newcommand{\vm}{\mathbf{m}}
\newcommand{\vn}{\mathbf{n}}
\newcommand{\vo}{\mathbf{o}}
\newcommand{\vp}{\mathbf{p}}
\newcommand{\vq}{\mathbf{q}}
\newcommand{\vr}{\mathbf{r}}
\newcommand{\Vs}{\mathbf{s}}
\newcommand{\vt}{\mathbf{t}}
\newcommand{\vu}{\mathbf{u}}
\newcommand{\vv}{\mathbf{v}}
\newcommand{\vw}{\mathbf{w}}
\newcommand{\vx}{\mathbf{x}}
\newcommand{\vy}{\mathbf{y}}
\newcommand{\vz}{\mathbf{z}}

\newcommand{\vone}{\mathbf{1}}
\newcommand{\vzero}{\mathbf{0}}

\newcommand{\valpha}{{\boldsymbol{\alpha}}}
\newcommand{\vbeta}{{\boldsymbol{\beta}}}
\newcommand{\vgamma}{{\boldsymbol{\gamma}}}
\newcommand{\vdelta}{{\boldsymbol{\delta}}}
\newcommand{\vepsilon}{{\boldsymbol{\epsilon}}}
\newcommand{\vzeta}{{\boldsymbol{\zeta}}}
\newcommand{\veta}{{\boldsymbol{\eta}}}
\newcommand{\vtheta}{{\boldsymbol{\theta}}}
\newcommand{\viota}{{\boldsymbol{\iota}}}
\newcommand{\vkappa}{{\boldsymbol{\kappa}}}
\newcommand{\vlambda}{{\boldsymbol{\lambda}}}
\newcommand{\vmu}{{\boldsymbol{\mu}}}
\newcommand{\vnu}{{\boldsymbol{\nu}}}
\newcommand{\vxi}{{\boldsymbol{\xi}}}
\newcommand{\vomikron}{{\boldsymbol{\omikron}}}
\newcommand{\vpi}{{\boldsymbol{\pi}}}
\newcommand{\vrho}{{\boldsymbol{\rho}}}
\newcommand{\vsigma}{{\boldsymbol{\sigma}}}
\newcommand{\vtau}{{\boldsymbol{\tau}}}
\newcommand{\vupsilon}{{\boldsymbol{\upsilon}}}
\newcommand{\vphi}{{\boldsymbol{\phi}}}
\newcommand{\vchi}{{\boldsymbol{\chi}}}
\newcommand{\vpsi}{{\boldsymbol{\psi}}}
\newcommand{\vomega}{{\boldsymbol{\omega}}}

\newcommand{\rLambda}{\mathrm{\Lambda}}
\newcommand{\rSigma}{\mathrm{\Sigma}}

\makeatletter
\DeclareRobustCommand\onedot{\futurelet\@let@token\@onedot}
\def\@onedot{\ifx\@let@token.\else.\null\fi\xspace}
\def\eg{\emph{e.g}\onedot} \def\Eg{\emph{E.g}\onedot}
\def\ie{\emph{i.e}\onedot} \def\Ie{\emph{I.e}\onedot}
\def\vs{\emph{vs\onedot}}
\def\cf{\emph{cf}\onedot} \def\Cf{\emph{C.f}\onedot}
\def\etc{\emph{etc}\onedot} \def\vs{\emph{vs}\onedot}
\def\wrt{w.r.t\onedot} \def\dof{d.o.f\onedot}
\def\etal{\emph{et al}\onedot}
\makeatother

\pgfplotstableread{
		l		cub
		1		37.66
		2		39.86
		4		40.48
		8		40.75
		16		40.62
	}{\augnumber}

\begin{abstract}
Deep metric learning for vision is trained by optimizing a representation network to map (non-)matching image pairs to (non-)similar representations. During testing, which typically corresponds to image retrieval, both database and query examples are processed by the same network to obtain the representation used for similarity estimation and ranking. In this work, we explore an asymmetric setup by light-weight processing of the query at a small image resolution to enable fast representation extraction. The goal is to obtain a network for database examples that is trained to operate on large resolution images and benefits from fine-grained image details, and a second network for query examples that operates on small resolution images but preserves a representation space aligned with that of the database network. We achieve this with a distillation approach that transfers knowledge from a fixed teacher network to a student via a loss that operates per image and solely relies on coupled augmentations without the use of any labels. In contrast to prior work that explores such asymmetry from the point of view of different network architectures, this work uses the same architecture but modifies the image resolution. We conclude that resolution asymmetry is a better way to optimize the performance/efficiency trade-off than architecture asymmetry. Evaluation is performed on three standard deep metric learning benchmarks, namely CUB200, Cars196, and SOP. Code: \url{https://github.com/pavelsuma/raml}
\end{abstract}

\section{Introduction}
\label{sec:intro}
\begin{figure}[t]
\vspace{-10pt}
  \centering
  \input{fig/cub_intro}
  \vspace{-3pt}
    \caption{Retrieval performance (mAP) \vs extraction cost of the query representation (GFLOPs) for the CUB200 dataset. The notation format used is ``database setup''$\rightarrow$``query setup'', where R50 and R18 are two variants of ResNet architecture. $M$ is equal to 448 and indicates the width and height of images. Contrary to the standard symmetric retrieval (circle), the query in the asymmetric setting is processed by a lighter network (triangle) or in a smaller resolution (diamond, pentagon).  \includegraphics[height=1.7ex]{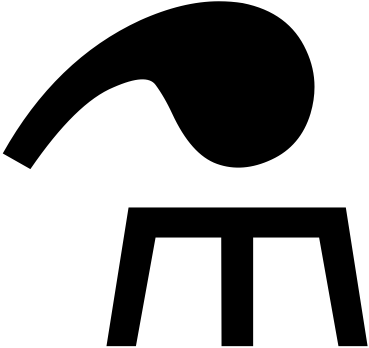}: networks trained with the proposed distillation approach to achieve resolution asymmetry (the focus of this work) and also network asymmetry for comparison.
\vspace{-5pt}
  \label{fig:cub_intro}
  }
\end{figure}

The performance of deep learning models typically increases with their size and computational complexity. Most work focuses on improving recognition performance and therefore relies on expensive models to train and deploy. Standard deep network architectures~\cite{hsc+19,hzr+16} are available in different variants that cover a range of the trade-off between performance and efficiency. Optimizing this trade-off forms a particular line of research that attracts a lot of attention since efficient and lightweight deep models allow deployment on mobile and low-resource devices or enable real-time execution. Therefore, powerful yet efficient models are desirable. One of the standard practices is to initially train a large model that is then used to obtain a smaller one, with weight pruning~\cite{hpt+15} and network distillation~\cite{hvd+15} being two dominant approaches to achieve that.

Network distillation uses the large model as a teacher that guides the training of a smaller student model. Most work in distillation is related to classification tasks, where the teacher logits act as a supervision~\cite{hvd+15}. Still, some work focuses on metric learning and image retrieval, where either the underlined vector representation~\cite{rbk+15} or pairwise scalar similarity values~\cite{pkl+19} function as supervision for distillation. In classification, once the small model is obtained, the large one is no longer used. However, due to the pairwise nature of the retrieval task, a possible asymmetry emerges; the query and database examples are processed by two different networks, with the network of the former being lightweight to reduce the extraction cost during query time. For small to medium size databases or with the use of fast nearest neighbor search methods~\cite{jds11,bal+16,ka14}, the extraction cost of the representation can be the test-time bottleneck. In the asymmetric setup, the two representation spaces corresponding to each of the two networks need to be aligned and compatible. This is the objective of asymmetric metric learning (AML) as introduced by Budnik and Avrithis~\cite{ba+21}.

AML is studied under the lens of asymmetric network architectures; the student model is a pruned variant of the teacher, a different but lighter architecture possibly discovered by neural architecture search. All these aspects reduce the query time. The resolution of the input images is an important aspect that is overlooked. The use of fully convolutional architectures allows any input resolution, while the representation extraction cost is roughly quadratic in that matter. Metric learning tasks that focus on instance-level recognition are known to benefit from the use of a large image resolution~\cite{rtc19,bjv+19}. The same holds for fine-grained recognition where object details matter~\cite{rms+20}, as also shown in this work. Therefore, input resolution is a critical parameter for the performance/efficiency trade-off. 

This work focuses on AML, where the asymmetry comes at the level of input resolution between the database network and the query network. The two architectures are the same, but the query network is trained at a low resolution to match the representation of the database network at a high resolution, which is performed with a distillation process that transfers knowledge from a teacher (database network) to the student (query network).
The contributions of this work are summarized as follows:
\vspace{-5pt}
\setlist[itemize]{align=parleft,left=0pt..1em}
\begin{itemize}
\setlength\itemsep{-5pt}
\item Asymmetry in the form of input image resolution is explored for the first time in deep metric learning.
\item A distillation approach is proposed to align the representation (absolute distillation) and the pairwise similarities (relational distillation) between student and teacher across task-tailored augmentations of the same image.
\item We conclude that resolution asymmetry is a better way to optimize the performance \vs efficiency trade-off compared to network asymmetry. 
\item As a side-effect, the student obtained with distillation noticeably outperforms the deep metric learning baselines in a conventional/symmetric retrieval.
\end{itemize}

A performance \vs efficiency comparison is shown in Figure~\ref{fig:cub_intro}. Compared to the baselines where a single network extracts both database and query examples with the same resolution (circle) and different resolution (diamond), distillation performs much better. The superiority of resolution over network asymmetry is evident too. More details about these experiments are discussed in Section~\ref{sec:exp}.
\section{Related work}
\label{sec:related}

\noindent\textbf{Asymmetric embedding compatibility.}
In image retrieval, embedding compatibility has to be ensured when the database examples are processed by a different network than the query examples. 
To this end, AML~\cite{ba+21} redefines standard metric learning losses in an asymmetric way, \ie the anchor example is processed by the query network, while the database network processes the corresponding positive and negative examples. However, for the objective of representation space alignment, these losses are outperformed by a simple unsupervised regression loss on the embeddings, a form of knowledge distillation between teacher and student. Using the supervised losses, on the other hand, boosts the performance of symmetric retrieval with the student, which even surpasses its teacher. Following the paradigm of unsupervised distillation, another recent approach compels the student to mimic the contextual similarities of image neighbors in the teacher embedding space~\cite{wwz+22}. Other than optimizing the weights of the student network, a generalization includes using neural architecture search to additionally optimize the network architecture~\cite{dzy+21} in a work that focuses on training for classification rather than in a metric learning manner.

Classification-based training is the dominant approach in a relevant task to ours, called backward compatible learning (BCT)~\cite{sxx+20}. However, the underlined task assumptions are different. Its objective is to add new data processed with a stronger backbone version without back-filling the current database. Compatibility is established with cross-entropy loss of the old classifier on old and new embeddings of the same input image.
This is extended to the compatibility of multiple embedding versions~\cite{hbc+22} or to tackling open-set backward compatibility with a continual learning approach~\cite{wcw+22}. Similarly, forward compatible learning~\cite{rvp+22} stores side information during training, which is leveraged in the future to transfer the old embeddings to another task.
Other methods for fixing inconsistent representation spaces include class prototypes alignment~\cite{bjw+21, zgs+22}, and transformation of both spaces, rather than a single one~\cite{wcy+20, hjl+19}.
\looseness=-1

Asymmetry also emerges when embeddings are collected from diverse devices which use different models, \eg in the domain of faces where the recognition should be compatible with all models~\cite{cwq+19}, or in localization and mapping task with multiple agents~\cite{dms+21}.

\noindent\textbf{Distillation and small image resolution. }
Image down-sampling remains the primary pre-processing step even at present times when the average visual memory of GPU allows processing bigger resolutions during training and testing. It is observed that using larger images reliably translates to higher performance regardless of the objective or dataset~\cite{rms+20}. Yet there are still many valid use cases where the inference has to be done with limited resources. 
In this context, distillation is used to align embeddings of high- and low-resolution images in the form of feature regression~\cite{gzl+20} or KL divergence~\cite{kaf+20}. 

Some networks are specifically trained to facilitate dynamic input resolution changes for an optimal speed-accuracy trade-off. Examples include distillation with mutual learning~\cite{yzc+20} or ensembling with the teacher learned on the fly~\cite{wsl+20}.
Distillation is also popular in the reverse direction going from small to large resolution, such as in the area of single image super-resolution~\cite{zcx+21, gzy+19, hdl+20, llk+20}.
\section{Method}
\label{sec:method}
Let $\cX$ be the space of all images, and $s: \cX \times \cX \rightarrow [-1,1]$ be a  similarity function that estimates scalar  similarity $s(x,q)$ for images $x,q\in \cX$, also called examples in the following. 

During testing, \ie for image retrieval, a similarity is estimated between the query example $q$ and each example in the database. Retrieval is then performed by ranking similarities in descending order.
In order to perform the retrieval efficiently, representation function $f: \cX \rightarrow \real^d$ is used that maps input examples from $\cX$ to a $d$-dimensional representation vector. These real-valued vectors, referred to as embeddings, are \l2-normalized and are used in combination with a standard similarity measure in $\real^d$ to obtain $s(\cdot)$. In this work, function $f(\cdot)$ is assumed to be implemented by a fully convolutional network (FCN), also denoted by $f_\theta(\cdot)$ declaring that the deep network is parametrized by parameter set $\theta$. As a result of using an FCN, the input image can be of any resolution. 

\subsection{Symmetric retrieval}
In the conventional task of symmetric retrieval, similarity $s(x,q)$ is computed by a simple dot product (equivalent to cosine similarity due to \l2-normalized embeddings) given by
\begin{equation}
    s(x,q) := s_s(x,q, f_\theta) = f_\theta(x)^T f_\theta(q),
\label{equ:ss}
\end{equation}
where representation function $f_\theta(\cdot)$ is used to process both the query and database examples in the same, symmetric, way, \ie same network architecture with the same parameters. Weights $\theta$ are optimized during the training phase according to semantic labels of pairs so that matching (non-matching) examples are mapped to nearby (faraway) embeddings in the representation space. 
This kind of training is the so called deep metric learning. Typical examples of losses are contrastive~\cite{hcl06}, triplet~\cite{wsl+14}, and multi-similarity~\cite{whh+19} that involve optimization of $s_s(\cdot)$ for training pairs\footnote{Some of the standard losses involve the use of Euclidean distance, but an alternative formulation with the use of similarity is possible. In this work, we employ similarities rather than distances.}. A network trained in such symmetric way constitutes the teacher in the following of this manuscript. Nonetheless, the teacher can equivalently be trained with other deep metric losses that do not directly involve pairwise comparison of train examples~\cite{tdt20,evt+20}. Both the testing and the training of the teacher are performed with the use of symmetric similarity  $s_s(\cdot)$. 

\subsection{Asymmetric retrieval - network-wise}
\label{sec:method_netasym}
This subsection provides background on the prior work which we rely on. 
Asymmetric similarity is defined as
\begin{equation}
    s(x,q) := s_{an}(x,q, f_\theta,g_\phi) = f_\theta(x)^T g_\phi(q),
\label{equ:san}
\end{equation}
where $g_{\phi}: \cX \rightarrow \real^d$ denotes a second FCN with parameter set $\phi$ to process the query only. The asymmetry is with respect to the network architectures processing query and database examples. The architecture of the query network $g(\cdot)$ is meant to be lighter than that of the database network $f(\cdot)$ to make the asymmetry meaningful in terms of query speed up. Following prior work~\cite{ba+21}, we assume that the database embeddings are fixed and extracted with $f_\theta(\cdot)$ which is trained by optimizing symmetric similarity; there is no option for modifying the database network or the database embeddings. 

Budnik and Avrithis~\cite{ba+21} perform a distillation process where knowledge from the fixed teacher (database) network is transferred to the student (query) network. This is performed by a regression process, also called absolute distillation, where the student embedding is optimized to match the teacher embedding for a particular training image $x$ such as by minimizing loss function $(1-s_{an}(x,x,f_\theta,g_\phi))^2$.
Their work concludes that this simple distillation process without the use of labels is the best performing approach for asymmetric retrieval as it directly reflects the objective of the task, \ie to align the two representation spaces. Combination with losses that use semantic labels performs worse for asymmetric retrieval as it compromises the alignment process.

\begin{table}[t]
\vspace{-5pt}
\centering
\newcommand*{\MinNumber}{0.0}\newcommand*{\MaxNumber}{0.0}\newcommand{\col}[1]{}
\small
\def\arraystretch{1.0}
\renewcommand*{\MinNumber}{10.9}
\renewcommand*{\MaxNumber}{43.9}
\definecolor{cub}{rgb}{0.298,0.686,0.314}
\definecolor{cars}{rgb}{1.0000,0.5961,0}
%
%
\renewcommand{\col}[1]{%
    \pgfmathsetmacro{\PercentColor}{100.0*(#1-\MinNumber+0.0*(\MaxNumber-\MinNumber))/(\MaxNumber-\MinNumber+0.0*(\MaxNumber-\MinNumber))}%
    \xdef\PercentColor{\PercentColor}%
    \cellcolor{cub!\PercentColor}{#1}%
}
\begin{tabular}{@{\ssp}r@{\ssp}||@{\ssp}r@{\ssp}|c@{\lsp}c@{\lsp}c@{\lsp}c@{\lsp}c@{\lsp}c|}
    \multicolumn{2}{c}{CUB200} & \multicolumn{6}{c}{test resolution} \\ \cmidrule(l{0pt}r{0pt}){3-8} \morecmidrules \cmidrule(l{0pt}r{0pt}){3-8}
    \multicolumn{2}{c}{} & 634 & 448 & 317 & 224 & 158 & \multicolumn{1}{c}{\hspace{-6pt}112} \\ \cline{3-8}
    \multirow{6}{*}{\rotatebox[origin=c]{90}{train resolution}}
    & 634& \col{43.9}& \col{42.2}& \col{37.3}& \col{30.3}& \col{22.7}& \col{15.3}\\ 
    & 448& \col{42.0}& \col{42.7}& \col{39.2}& \col{32.8}& \col{25.0}& \col{17.5}\\ 
    & 317& \col{35.9}& \col{40.7}& \col{39.6}& \col{34.1}& \col{26.3}& \col{18.3}\\ 
    & 224& \col{21.6}& \col{31.6}& \col{36.3}& \col{34.3}& \col{27.9}& \col{20.7}\\ 
    & 158& \col{11.4}& \col{19.1}& \col{27.5}& \col{31.7}& \col{28.3}& \col{21.9}\\ 
    & 112& \col{10.9}& \col{16.8}& \col{22.9}& \col{28.0}& \col{26.8}& \col{21.9}\\ 
    \cline{3-8} 
\end{tabular}\\[5pt]
\small
\def\arraystretch{1.0}
\renewcommand*{\MinNumber}{5.0}
\renewcommand*{\MaxNumber}{42.3}
\definecolor{cub}{rgb}{0.298,0.686,0.314}
\definecolor{cars}{rgb}{1.0000,0.5961,0}
%
%
\renewcommand{\col}[1]{%
    \pgfmathsetmacro{\PercentColor}{100.0*(#1-\MinNumber+0.0*(\MaxNumber-\MinNumber))/(\MaxNumber-\MinNumber+0.0*(\MaxNumber-\MinNumber))}%
    \xdef\PercentColor{\PercentColor}%
    \cellcolor{cub!\PercentColor}{#1}%
}
\begin{tabular}{@{\ssp}r@{\ssp}||@{\ssp}r@{\ssp}|c@{\lsp}c@{\lsp}c@{\lsp}c@{\lsp}c@{\lsp}c|}
    \multicolumn{2}{c}{Cars196} & \multicolumn{6}{c}{test resolution} \\ \cmidrule(l{0pt}r{0pt}){3-8} \morecmidrules \cmidrule(l{0pt}r{0pt}){3-8}
    \multicolumn{2}{c}{} & 634 & 448 & 317 & 224 & 158 & \multicolumn{1}{c}{\hspace{-6pt}112} \\ \cline{3-8}
    \multirow{6}{*}{\rotatebox[origin=c]{90}{train resolution}}
    & 634& \col{40.4}& \col{35.5}& \col{26.7}& \col{16.7}& \col{9.3}& \col{5.0}\\
    & 448& \col{42.3}& \col{41.6}& \col{34.5}& \col{23.0}& \col{13.0}& \col{6.3}\\
    & 317& \col{31.9}& \col{36.2}& \col{33.8}& \col{25.2}& \col{15.4}& \col{7.6}\\
    & 224& \col{22.8}& \col{31.3}& \col{33.3}& \col{29.3}& \col{20.7}& \col{10.9}\\
    & 158& \col{12.7}& \col{21.0}& \col{27.3}& \col{29.1}& \col{25.2}& \col{15.4}\\
    & 112& \col{6.0}& \col{10.4}& \col{16.7}& \col{22.1}& \col{22.1}& \col{16.6}\\
    \cline{3-8} 
\end{tabular}
\\[5pt]
\small
\def\arraystretch{1.0}
\renewcommand*{\MinNumber}{30.0}
\renewcommand*{\MaxNumber}{61.6}
\definecolor{cub}{rgb}{0.298,0.686,0.314}
\definecolor{cars}{rgb}{1.0000,0.5961,0}
%
%
\renewcommand{\col}[1]{%
    \pgfmathsetmacro{\PercentColor}{100.0*(#1-\MinNumber+0.0*(\MaxNumber-\MinNumber))/(\MaxNumber-\MinNumber+0.0*(\MaxNumber-\MinNumber))}%
    \xdef\PercentColor{\PercentColor}%
    \cellcolor{cub!\PercentColor}{#1}%
}
\begin{tabular}{@{\ssp}r@{\ssp}||@{\ssp}r@{\ssp}|c@{\lsp}c@{\lsp}c@{\lsp}c@{\lsp}c@{\lsp}c|}
    \multicolumn{2}{c}{SOP} & \multicolumn{6}{c}{test resolution} \\ \cmidrule(l{0pt}r{0pt}){3-8} \morecmidrules \cmidrule(l{0pt}r{0pt}){3-8}
    \multicolumn{2}{c}{} & 634 & 448 & 317 & 224 & 158 & \multicolumn{1}{c}{\hspace{-6pt}112} \\ \cline{3-8}
    \multirow{6}{*}{\rotatebox[origin=c]{90}{train resolution}}
    & 634& \col{60.7}& \col{58.1}& \col{52.9}& \col{45.1}& \col{37.4}& \col{30.0}\\
    & 448& \col{61.0}& \col{61.6}& \col{57.5}& \col{50.8}& \col{42.4}& \col{33.9}\\
    & 317& \col{59.1}& \col{59.4}& \col{60.4}& \col{51.8}& \col{43.7}& \col{34.1}\\
    & 224& \col{55.9}& \col{60.0}& \col{60.0}& \col{57.4}& \col{50.8}& \col{41.0}\\
    & 158& \col{51.3}& \col{55.2}& \col{55.9}& \col{55.2}& \col{53.2}& \col{41.5}\\
    & 112& \col{45.8}& \col{49.6}& \col{51.8}& \col{51.9}& \col{48.7}& \col{47.2}\\
    \cline{3-8} 
\end{tabular}
\\
  \vspace{2pt}
  \caption{Retrieval performance with the use of symmetric similarity when the network is trained and tested on different image resolution. This experiment does not include any asymmetry between database and query. ResNet-50 is used as the backbone and trained with labels and triplet loss (equivalent to the teacher network in our approach). Mean Average Precision is reported.\label{tab:teacher_sizes} 
\vspace{-5pt}
}
\end{table}

\subsection{Asymmetric retrieval - resolution-wise}
In this subsection, we first highlight the impact of image resolution for symmetric similarity and then introduce resolution asymmetry.

Even though the input resolution can be arbitrary, the level of details seen during training influences the ability to capture information in the representation. 
In practice, we make the following two observations:
(i) the resolution used in training imposes restrictions on the resolution used during test time~\cite{tvd+19}; we presume the network gets adjusted to a specific level of details and scale of objects or their parts. (ii) fine-grained recognition, which is the main focus of this work, benefits from larger resolutions than the ones typically used for image classification tasks. 
Results in Table~\ref{tab:teacher_sizes} support these observations; performance declines when there is a test-train resolution discrepancy. An exception appears for small training resolution where asymmetry with a slightly larger test resolution performs better; the benefit of the larger train image resolution is higher than the harm of the asymmetry.
As shown in the work of Berman \etal~\cite{bjv+19}, the mentioned discrepancy can be alleviated by re-parametrization of a non-linear pooling operation, but this does not apply to the asymmetric task of this work.

The asymmetry in Section~\ref{sec:method_netasym} is caused by using two different network architectures to process query and database examples. This work introduces an asymmetry in the resolution of each model's input images.
In that case, the asymmetric similarity is firstly defined in a simplistic way given by
\begin{equation}
s(x,q):= s_{ar}(x,q, f_\theta) = f_\theta(x)^T f_\theta(r(q)),
\label{equ:sarstar}
\end{equation}
where $r:\cX \rightarrow \cX$ is an image down-sampling function. 
In the case of network asymmetry, two different networks cannot be used without any training for alignment of the representation spaces. Conversely, in the case of resolution asymmetry, the same network with the same parameters can be used, as in \equ{sarstar}, at different resolutions. Therefore, embeddings of two different resolutions are matched to each other despite the discrepancy in the average object scale, which is expected to harm performance compared to the symmetric case. 

We go one step further from \equ{sarstar} and re-parametrize the query network and define asymmetric similarity by
\begin{equation}
s(x,q):= s_{ar}(x,q, f_\theta, f_\phi) = f_\theta(x)^T f_\phi(r(q)),
\label{equ:sar}
\end{equation}
where the architecture for the database and query networks are identical, but their parameters differ. In the following, we discuss how to optimize $\phi$ for resolution-wise asymmetric similarity.

\begin{figure*}[t]
  \vspace{-8pt}
  \centering
  \includegraphics[width=0.9\textwidth]{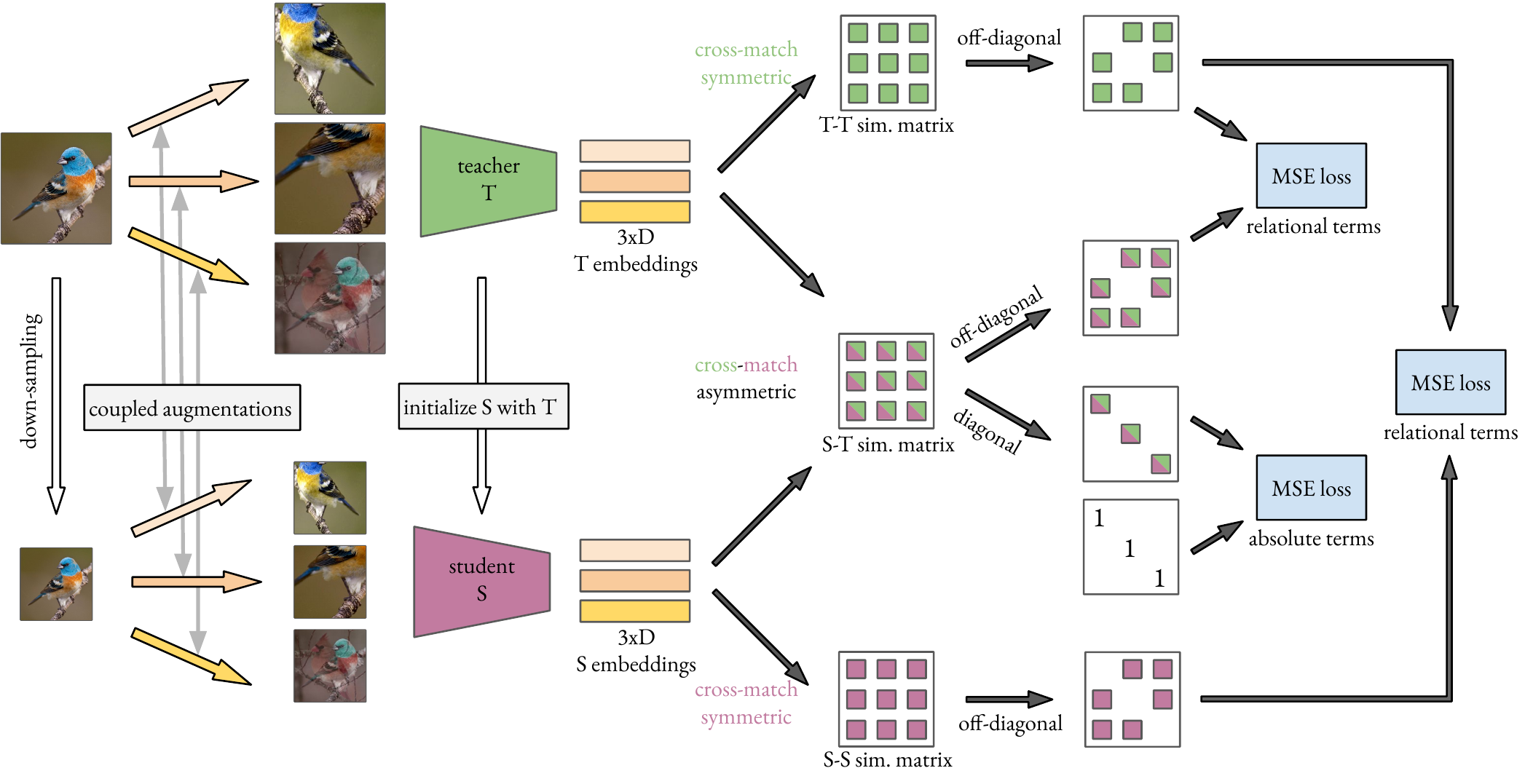}
  \vspace{-5pt}
  \caption{An overview of the proposed approach. The teacher network, which processes database images during testing, is pre-trained to operate on large image resolution and is now fixed and used to distill knowledge to the student network that operates on small image resolution which processes queries during testing. The similarity is computed between all embeddings of different augmentations of the same image with the same network and across networks. The resulting similarity matrices are used in three distillation losses during training. MSE: mean squared error.
  \label{fig:teaser}
  \vspace{-8pt}
  }
\end{figure*}

\paragraph{Training.}
The teacher model $f_\theta$ is given and is pre-trained on the training examples at the large resolution.
Teacher parameters are frozen for the entirety of our training.
We initialize the student $f_\phi$ with the parameters $\theta$ of the teacher; this is a good initialization aligning the representation spaces up to the resolution discrepancy. Note that there is no such good initialization for network-wise asymmetry. 
We use two types of losses, one to perform absolute distillation and one to perform relational distillation~\cite{rmo+21,pkl+19}. 
A visual overview of the proposed method is given in Figure~\ref{fig:teaser}.
We also refer to teacher and student networks by $T(\cdot)$ and $S(\cdot)$ for brevity, where the latter already includes the down-sampling process.

\emph{Absolute distillation} is performed, as in the original AML work~\cite{ba+21},  by regression between the teacher and student outputs given by loss
\begin{equation}
\ell_{abs}(x, \theta; \phi) = \Bigl(1 - s_{ar}(x,x,f_\theta, f_\phi)\Bigr)^2 .
\label{equ:abs1}
\end{equation}
Note that $\ell_{abs}$ is bounded between 0 and 4.
This loss does not involve any labels and is applicable to any example in the training set, allowing for alignment of the representation space at a large number of training embeddings.

\emph{Coupled augmentations. }
A common trick to introduce curated noise is to randomly alter each network input by handcrafted, domain-specific functions, \ie random image augmentations. During supervised training, the model learns to be invariant to such noise since the alterations of a single image correspond to the same label. Our proposed method is fully unsupervised but utilizes random augmentations to virtually increase the training set and the number of training embeddings used for alignment. We empirically observe that it is essential to randomly perturb the input in the same way for both teacher and student involved in the asymmetric similarity. 
These are what we refer to as coupled augmentations in this work. 
Knowledge distillation is known to be highly compatible and greatly benefit from data augmentations~\cite{wlj+20}; this is observed in our case too but only with coupled augmentations. Experiments and discussion for coupled and non-coupled augmentations are included in Section~\ref{sec:exp}.

More formally, we define $a(x)$ as the set of examples obtained by applying different image augmentations to $x$. Then, the average absolute distillation loss\footnote{In the mathematical formulation, we first augment and then down-sample, which is a different order than in Figure~\ref{fig:teaser}, but identical in practice.} is given by
\begin{equation}
\cL_{abs}(x, a, \theta; \phi) = \frac{1}{|a(x)|} \sum_{z \in a(x)} \ell_{abs}(z, \theta; \phi).
\label{equ:abs}
\end{equation}

\emph{Relational distillation} is performed by using scalar similarities, instead of the embeddings, to guide the student. 
What is preserved is the relative comparison between two examples, therefore the name relational~\cite{pkl+19}. 
The two examples involved are different augmentations of the same original example, resulting in a loss being applied per original example separately. We propose two alternatives where the teacher-to-teacher relation is distilled to a teacher-to-student relation or student-to-student relation. 
The former is given by
\begin{equation}
\cL_{rel-ts}(x, a, \theta; \phi) = \frac{1}{n} \displaystyle\sum_{\substack{y,z \in a(x)\\z\neq y}} \ell_{rel-ts}(y,z,\theta; \phi)\\,
\label{equ:relallts}
\end{equation}
where $n=|a(x)|^2-|a(x)|$ and
\begin{eqnarray}
\ell_{rel-ts}(y,z, \theta; \phi) =&\hspace{-10pt} \Bigl( s_s(y,z, f_\theta) - s_{ar}(y,z, f_\theta, f_\phi) \Bigr)^2\nonumber\\
=&\hspace{-10pt} \Bigl( f_\theta(y)^\top f_\theta(z) - f_\theta(y)^\top f_\phi(r(z)) \Bigr)^2\nonumber\\
                =&\hspace{-15pt} \Bigl( T(y)^\top T(z) - T(y)^\top S(z) \Bigr)^2.
\label{equ:relonets}
\end{eqnarray}
Equivalently to $\cL_{rel-ts}$ in \equ{relallts}, we use $\cL_{rel-ss}$ by defining
\begin{align}
\ell_{rel-ss}(y,z, \theta; \phi) =& \hspace{4pt} \Bigl( s_s(y,z, f_\theta) - s_s(r(y),r(z), f_\phi) \Bigr)^2\nonumber\\
=& \Bigl( f_\theta(y)^\top f_\theta(z) \hspace{-2pt}-\hspace{-2pt} f_\phi(r(y))^\top f_\phi(r(z)) \Bigr)^2\nonumber\\
                =&\hspace{8pt} \Bigl( T(y)^\top T(z) - S(y)^\top S(z) \Bigr)^2. \label{equ:reloness}
\end{align}
Relational distillation in the form of \equ{reloness} is shown to fail in prior work for asymmetric retrieval~\cite{ba+21} as it does not satisfy the alignment objective, while in the form of \equ{relonets} is shown effective~\cite{wwz+22}. In our case, such distillation loss terms are effective even by themselves under particular conditions, which is shown and discussed in Section~\ref{sec:exp}. Note that one of our relational terms, \ie \equ{relonets} is similar to that in the work of Wu \etal~\cite{wwz+22}. Nevertheless, we do not require a costly nearest neighbors process and simply use random augmentations.

Each of the three losses is averaged over all examples in a batch, resulting to $\cL_{abs}$, $\cL_{rel-ts}$, and $\cL_{res-ss}$, respectively. The the total loss is given by
\begin{equation}
\vspace{7pt}
    \cL = \cL_{abs} + \lambda_{t}\cL_{rel-ts} + \lambda_{s}\cL_{rel-ss},
\label{equ:l_final}
\end{equation}
where $\lambda_t$ and $\lambda_s$ are hyper-parameters tuned during cross-validation.
\section{Experiments}
\label{sec:exp}
In this section, we provide the implementation details and present the experimental results.

\subsection{Datasets}
We use three standard deep metric learning datasets for fine-grained recognition, namely Caltech-UCSD Birds (\textit{CUB200}) dataset~\cite{wbm+10},  Stanford Cars dataset (\textit{Cars196})~\cite{ksd+13}, and  Stanford Online Products (\textit{SOP})~\cite{sxj+15}. They comprise 11,788 images of birds from 200 classes,  16,185 images of cars from 196 classes, and 120,053 images of products from 22,634 classes, respectively. Following common practice, for CUB200 and Cars196 we use the first half of the classes for training and the other half for testing, while for SOP we use the provided train/test sets.

\subsection{Experimental setup}
As the first step, we use the labeled training set to train the teacher. All training images are re-sampled to have their largest dimension equal to what we refer to as database resolution or large resolution. Then, the teacher weights are used to initialize the student, while we freeze the teacher weights and optimize the student with the proposed distillation losses on the whole training set. During this process, the student receives input images at what we refer to as query resolution or small resolution. We perform the student training three times with different, but fixed, seeds.
In the evaluation phase, performance is evaluated by mean Average Precision (mAP) and by recall at 1 (R@1) which is equal to 1 if the top retrieved image is from the correct class. We report average performance across seeds.

The student network is evaluated in two ways. Firstly, for asymmetric retrieval where the database (query) examples are processed by the teacher (student) network in the large (small) resolution. Secondly, we drop the assumption that the database embeddings are fixed and evaluate the student network in a symmetric manner where it processes both database and query examples in the small resolution.
The teacher network, other than participating in the distillation process and in the aforementioned asymmetric testing for retrieval, is also evaluated in the following two ways to provide a baseline. Firstly, it is evaluated in a symmetric way by processing both database and query examples at the resolution it was trained with. Secondly, it is evaluated in an asymmetric way by processing database (query) examples at the large (small) resolution (a single network is processing images at two different resolutions). 
To summarize, our evaluation setups are \emph{teacher-student asymmetric} (two networks, two resolutions, same architecture) by \equ{sar}, \emph{student symmetric} by \equ{ss}, \emph{teacher symmetric} by \equ{ss}, \emph{teacher asymmetric} (one network, two resolutions) by \equ{sarstar}. 

\subsection{Implementation details}
ResNet-50 (R50) and ResNet-18 (R18)~\cite{hzr+16} are used as backbone FCN, with initial weights obtained from ImageNet~\cite{dsl+09} pre-training. Generalized mean pooling (GeM)~\cite{rtc19} and a fully connected layer reducing the final embedding dimension to 512 are added on top of the FCN. Optimization is performed with AdamW~\cite{lh+19} and a one cycle learning rate scheduler~\cite{st+17} using PyTorch default values.
Teacher networks are trained with triplet loss and distance weighed negative mining~\cite{wms+17}. Following standard protocol~\cite{rms+20,bxk+20,chx+19}, we perform random crop~\cite{slj+15}, resize to fixed resolution and flip horizontally with probability $0.5$. The number of epochs is set to 200. Batch size equals 200 for CUB200 and Cars196 and 4000 for SOP.

During student training with distillation, the same augmentation strategy is used with the addition of color distortion with the strength of 0.5~\cite{ckn+20} and image mixup~\cite{zcd+18} with $\alpha=0.2$. For mixup, each image is mixed with the next image in the batch.
We train for 200 epochs with batch size equal to 200, limit the number of examples to 8000 per epoch, and use 8 different augmentations per image unless otherwise stated.
For proper hyper-parameter tuning, we use Optuna library~\cite{asy+19} and half of the training set as validation set~\cite{tdt20}. We tune learning rate, $\lambda_t$, and $\lambda_s$; indicative values are 1.1e-4, 0.7, and 0.7, respectively.

The evaluation policy is identical for all experiments.
The input image is re-sampled to the large or small resolution depending on the setup and network used while preserving the aspect ratio.
Following common practice, the center square area is cropped.

\subsection{Results}

\textbf{Performance comparison at multiple query resolutions. }
In Table~\ref{tab:results} we show the performance of our distillation method and compare it with baselines on three smaller image resolutions. Compared with the respective (same query resolution) baselines for symmetric retrieval (1st block), the distilled students (3rd block) are on all datasets higher in terms of their asymmetric retrieval performance. The benefit of asymmetry and using large resolution database images gets larger for decreasing query resolution (increasing query extraction savings). We also substantially improve over the naive asymmetric approach (2nd block \vs 3rd block), where the teacher network processes the small resolution queries without any appropriate training.

If the assumption of fixed database embeddings is dropped, it makes sense to look at the student symmetric retrieval performance at small resolutions; database extraction cost is also reduced now.  
First, we observe that through distillation, symmetric retrieval performance (4th block) is much higher than training with the standard deep metric learning way (1st block). Note that the same is achieved in earlier work on asymmetric metric learning, but only with the use of labels~\cite{ba+21}, while we do not use labels for student training. Secondly, we observe that the student's symmetric performance (measured via mAP) is larger than its asymmetric one if the query resolution is relatively large. This does not hold for smaller query resolutions where asymmetry is still more meaningful. Focusing on recall at 1, the student performs in most cases higher in the symmetric setting than in the asymmetric, indicating a different behavior for the top-ranked examples; the asymmetric setup works better considering all relevant examples but worse if we consider the most similar ones.\looseness=-1

\begin{table}[t]
\vspace{-5pt}
  \centering
\small
\def\arraystretch{0.8}
  \begin{tabular}{l@{\ssp}l@{\ssp}c@{\ssp}c@{\msp} c@{\msp}c@{\lsp}c@{\msp}c@{\lsp}c@{\msp}c }
    \toprule
    &&& & \multicolumn{2}{c}{CUB200} & \multicolumn{2}{c}{Cars196} & \multicolumn{2}{c}{SOP}\\\midrule
    QR & DR  & QN & DN & mAP & R@1 & mAP & R@1 & mAP & R@1 \\\midrule
    \multicolumn{10}{c}{\emph{teacher symmetric:}}\\
    \multicolumn{10}{c}{\footnotesize teacher trained \& tested @ DR, different network per row}\\\midrule
    448 & 448 & T & T & 42.7 & 73.7 & 41.6 & 87.9 & 62.6 & 82.1 \\
    317 & 317 & T & T & 39.6 & 71.6 & 33.8 & 81.4 & 60.4 & 81.2 \\
    224 & 224 & T & T & 34.3 & 64.6 & 29.6 & 75.2 & 57.4 & 79.2 \\
    158 & 158 & T & T & 28.3 & 55.9 & 25.2 & 64.1 & 53.2 & 76.0 \\\midrule
    \multicolumn{10}{c}{\emph{teacher asymmetric:}}\\
    \multicolumn{10}{c}{\footnotesize  teacher trained @ 448, same network for all rows}\\\midrule
    317 & 448 & T & T & 39.9 & 69.5 & 36.5 & 81.3 & 58.4 & 80.1 \\
    224 & 448 & T & T & 34.3 & 61.0 & 25.7 & 61.0 & 51.3 & 74.3 \\
    158 & 448 & T & T & 26.6 & 47.6 & 14.8 & 32.3 & 38.2 & 60.1 \\\midrule
    \multicolumn{10}{c}{\emph{teacher-student asymmetric:}}\\
    \multicolumn{10}{c}{\footnotesize teacher trained @ 448, student trained with distillation @ QR}\\
    \multicolumn{10}{c}{\footnotesize  same teacher for all rows, different student per row}\\\midrule
    317 & 448 & S & T& 42.3 & 72.9  & 41.0 & 87.1 &  61.3 & 81.9  \\
    224 & 448 & S & T& 40.8 & 70.5  & 38.1 & 83.4 &  59.7 & 80.8  \\
    158 & 448 & S & T& 36.9 & 63.7  & 31.8 & 71.3 &  56.4 & 78.1  \\\midrule
    \multicolumn{10}{c}{\emph{student symmetric:}}\\
    \multicolumn{10}{c}{\footnotesize student trained with distillation @ QR, different student per row}\\ \midrule
    317 & 317  & S & S & 43.6 & 74.9  & 41.8 & 88.1 &  61.6 & 82.0 \\
    224 & 224  & S & S & 40.9 & 71.8  & 38.1 & 85.0 &  59.5 & 80.8 \\
    158 & 158  & S & S & 36.0 & 66.8  & 30.6 & 76.3 &  55.3 & 77.9 \\
    \bottomrule
  \end{tabular}  
  \caption{Performance results for resolution-wise asymmetric and symmetric retrieval at different query resolution. 
  QR,DR: query and database resolution. QN,DN: query and database network. S, T: student and teacher. 
  \label{tab:results}
  \vspace{-5pt}
  }
\end{table}

\textbf{Ablation study. }
Our work heavily relies on the use of augmentations. We dissect their contribution in Table~\ref{tab:aug_ablation}. They are organized into three groups: \textit{coupled} applies the same image transformations in the teacher and student (see Figure~\ref{fig:teaser}), \textit{geometric augmentations} correspond to random resized crop, color jitter, and horizontal flip, and \textit{MX} constitutes image mixup. The setup with augmentations performed only separately in the teacher input and student input is clearly worse and even harms the alignment. Such non-coupled augmentations can potentially increase the invariance of the asymmetric similarity; its failure signifies that representation space alignment is the important objective and not invariance, which is inherited from the teacher anyway.
Mixup by itself is not as good as the more standard augmentations but still improves the final method. It seems that non-standard augmentations are effective if applied in a coupled way. \looseness=-1

We additionally show the performance of each loss term from~\equ{l_final} separately along with their combination. 
We confirm that absolute distillation by itself is already a strong performer for our asymmetric resolution setting. It outperforms the other two relative loss terms. Nevertheless, their combination is the top-performing approach. Note that the relational loss terms have negligible added cost in the training since all the embeddings are already obtained for the needs of the absolute distillation loss term.
To our surprise, $\cL_{rel-ss}$ by itself improves the asymmetric performance, although the individual loss serves only as a regular relational knowledge distillation previously shown to fail for asymmetric retrieval~\cite{ba+21}. We investigate this in the next experiment.
\looseness=-1

\begin{table}[t]
  \vspace{-5pt} 
  \centering
  \small
\def\arraystretch{0.87}
  \begin{tabular}{@{\msp}l@{\msp}c@{\msp}c@{\msp}c@{\msp}c@{\msp}c@{\msp}c@{\msp}c@{\msp}}
    \toprule
    \multirow{2}{*}{loss} & \multirow{2}{*}{coupled} & \multirow{2}{*}{G} & \multirow{2}{*}{MX} & \multicolumn{2}{c}{\hspace{-5pt}\textit{asymmetric}} & \multicolumn{2}{c}{\hspace{-5pt}\textit{symmetric}} \\
    & & & & mAP & R@1 & mAP & R@1 \\
    \midrule
    $\cL_{abs}$ & & & & 37.7 & 66.1 & 34.8 & 65.1 \\
    $\cL_{abs}$ & & \checkmark & & 31.3 & 55.6 & 30.7 & 60.3 \\
    $\cL_{abs}$ & & & \checkmark & 20.9 & 40.1 & 29.8 & 60.0 \\
    \midrule
    $\cL_{abs}$ & \checkmark & \checkmark & & 40.1 & 69.8 & 40.6 & 71.5 \\
    $\cL_{abs}$ & \checkmark & & \checkmark & 39.5 & 68.5 & 38.8 & 68.9 \\
    $\cL_{abs}$ & \checkmark & \checkmark & \checkmark & 40.3 & 69.8 & 40.5 & 71.2 \\
    \midrule
    $\cL_{rel-ts}$ & \checkmark & \checkmark & \checkmark & 39.7 & 68.9 & 39.1 & 69.2 \\
    $\cL_{rel-ss}$ & \checkmark & \checkmark & \checkmark & 37.9 & 65.1 & 40.0 & 71.4 \\
    \midrule
    $\cL$ & \checkmark & \checkmark & \checkmark & 40.8 & 70.5 & 40.9 & 71.8 \\
    \bottomrule
  \end{tabular}

  \caption{Impact of the loss function and of different augmentation strategies on performance for teacher-student asymmetric and student symmetric retrieval after distillation. Results reported on CUB200. G: geometric augmentations. MX: mixup at the input level. Teacher (student) operates at 448 (224) resolution.\label{tab:aug_ablation}
  \vspace{-5pt}   
  }
\end{table}

\textbf{Impact of student initialization. }
If we initialize the student network with the teacher weights, an initial alignment of the two embedding spaces is provided, which plays a crucial role, as seen in Table~\ref{tab:inits_ablation}.
Note that no such initialization exists in the case of network-wise asymmetry. It is possible in our case because it corresponds to matching objects at different resolutions with the same network, which is a good starting point that is then improved via distillation.
We compare such initialization to the one using weights from a teacher trained for the small resolution and the one with ImageNet weights.
Both alternatives perform worse than the suggested and are on par with each other. Even using each of the losses separately performs much better given a good initialization. We suspect the reason is that the embedding spaces in these two cases are initially entirely misaligned.

Noticeably, the relational distillation with $\cL_{rel-ss}$, which does not involve any asymmetric term, completely fails without this initialization but performs well with it. We presume that such a relational distillation only works if we are already close to a good solution. Otherwise, satisfying its objective does not meet the representation space alignment objective at all.

\begin{table}[t]
  \vspace{-8pt}
  \centering
\small
\def\arraystretch{0.8}
  \begin{tabular}{llrrrr}
    \toprule
    \multirow{2}{*}{initialization} & \multirow{2}{*}{loss} & \multicolumn{2}{c}{\hspace{-5pt}\textit{asymmetric}} & \multicolumn{2}{c}{\hspace{-5pt}\textit{symmetric}} \\
    & & mAP & R@1 & mAP & R@1 \\
    \midrule
    teacher@448 & $\cL$ & 40.8 & 70.5 & 40.9 & 71.8  \\
    teacher@224 & $\cL$ & 36.3 & 62.9 & 37.0 & 66.8 \\
    ImageNet  & $\cL$ & 36.3 & 62.4 & 36.9 & 67.3 \\
    \midrule
    teacher@224 & $\cL_{abs}$ & 35.7 & 61.6 & 35.9 & 65.5 \\
    teacher@224 & $\cL_{rel-ts}$ & 36.3 & 60.8 & 38.9 & 68.5  \\
    teacher@224 & $\cL_{rel-ss}$ & 1.3 & 1.1 & 39.6 & 69.8 \\
    \midrule
    teacher@448 & $\cL_{abs}$ & 40.3 & 69.8 & 40.5 & 71.2 \\
    teacher@448 & $\cL_{rel-ts}$ &  39.7 & 68.9 & 39.1 & 69.2 \\
    teacher@448 & $\cL_{rel-ss}$ & 37.9 & 65.1 & 40.0 & 71.4 \\
    \bottomrule
  \end{tabular}
  \caption{Impact of the student initialization during our distillation process with the different losses used all together or separately. Performance is evaluated on CUB200 for teacher-student asymmetric retrieval and for student symmetric retrieval. Initialization is performed by the teacher trained at the large or small resolution, or with ImageNet pre-trained weights.
  Teacher (student) operates at 448 (224) resolution.
   \label{tab:inits_ablation}
  }
\end{table}

\textbf{Impact of number of augmentations.}
The amount of terms in the relative distillation loss grows quadratically with the number of augmentations per image, while training time increases more or less linearly. Figure~\ref{fig:aug_number} shows a noticeable increase in performance when using more than one augmentation of the same image. The gain saturates after 8, which is the number of augmentations we use in the rest of the experiments. \looseness=-1

\begin{figure}[t]
  \centering
  \begin{tikzpicture}
\begin{axis}[%
	width=0.88\linewidth,
	height=0.44\linewidth,
	xlabel={\normalsize number of augmentations per image},
	ylabel={\normalsize mAP},
    ymax = 41,
    ymin = 37,
    xmin = 0,
    xmax = 17,
    grid=both,
    xtick={1,2,4,8,16},
    ytick={37,38,...,41},
]

	\addplot[color=blue,     solid, mark=*,  mark size=1.5, line width=1.0] table[x=l, y expr={\thisrow{cub}}]  \augnumber;
\end{axis}
\end{tikzpicture}
  \caption{Impact of the number of augmentations on the performance of teacher-student asymmetric retrieval after our distillation on CUB200.
  \label{fig:aug_number}}
\end{figure}

\begin{figure}[t]
\vspace{-15pt}
  \centering
  \pgfmathsetmacro{\teasermarkersize}{2.5}

\begin{tikzpicture}
\begin{axis}[%
  width=0.98\linewidth,
  height=0.75\linewidth,
  ylabel={\small performance (mAP)},
  xlabel={\small  query extraction cost (GFLOPs)},
  legend pos=south east,
  tick label style={font=\small},
  ylabel near ticks, xlabel near ticks, 
  legend style={font=\scriptsize}, 
  xmin = -1, xmax = 18,ymin = 13, ymax = 45,
  xlabel style={yshift=1ex},
  grid=minor,
  ]
    \addplot[color=\symcol, mark=\bothmark,mark size=\teasermarkersize pt, only marks] coordinates {(16.439,41.61)};
    \node [above] at (axis cs:  16.439,41.61) {\textcolor{\symcol}{\tiny R50:M}};
    \addplot[color=\symcol, mark=\bothmark,mark size=\teasermarkersize pt, only marks] coordinates {(4.110,29.25)};
    \node [above] at (axis cs: 4.110,29.25) {\textcolor{\symcol}{\tiny R50:$0.5$M}};
    \addplot[color=\symcol, mark=\bothmark,mark size=\teasermarkersize pt, only marks] coordinates {(7.274,34.64)};
    \node [left] at (axis cs:  7.274,34.64) {\textcolor{\symcol}{\tiny R18:M}};
    \addplot[color=\symcol, mark=\bothmark,mark size=\teasermarkersize pt, only marks] coordinates {(1.819,26.23)};
    \node [above] at (axis cs: 1.819,26.23) {\textcolor{\symcol}{\tiny R18:$0.5$M}};
%
%
%
%
    \addplot[color=\asmrescol, mark=\resmark,mark size=\teasermarkersize pt, only marks] coordinates {(4.110,25.7)};
    \node [below] at (axis cs:  4.110,25.7) {\textcolor{\asmrescol}{\tiny R50:$M$\hspace{-0.5pt}$\rightarrow\hspace{-2.5pt}0.5M$}};
    \addplot[color=\asmrescol, mark=\resmark,mark size=\teasermarkersize pt, only marks] coordinates {(8.384,36.5)};
    \node [above] at (axis cs:  8.384,36.5) {\textcolor{\asmrescol}{\tiny R50:$M$\hspace{-0.5pt}$\rightarrow\hspace{-2.5pt}0.7M$}};
    \addplot[color=\asmrescol, mark=\resmark,mark size=\teasermarkersize pt, only marks] coordinates {(2.096,14.8)};
    \node [above] at (axis cs:  2.096,14.8) {\textcolor{\asmrescol}{\tiny R50:$M$\hspace{-0.5pt}$\rightarrow\hspace{-2.5pt}0.35M$}};
%
%
%
\addplot[color=\asmnetcol, mark=\netmark,mark size=\teasermarkersize pt, only marks] coordinates {(7.274,32.36)};
    \node [below] at (axis cs: 7.274,32.36) {\textcolor{\asmnetcol}{\tiny R50\hspace{-0.5pt}$\rightarrow$\hspace{-.5pt}R18:$M$\cite{ba+21}} };
%
%
%

\addplot[color=\asmdrescol, mark=\dresmark,mark size=\teasermarkersize pt, only marks] coordinates {(4.110,38.13)};
    \node [above] at (axis cs:  4.110,38.13) {\textcolor{\asmdrescol}{\tiny R50:$M$\hspace{-0.5pt}$\rightarrow\hspace{-2.5pt}0.5M$}};
\addplot[color=\asmdrescol, mark=\dresmark,mark size=\teasermarkersize pt, only marks] coordinates {(8.384,40.99)};
    \node [above] at (axis cs:  8.384,40.99) {\textcolor{\asmdrescol}{\tiny R50:$M$\hspace{-0.5pt}$\rightarrow\hspace{-2.5pt}0.7M$}};
\addplot[color=\asmdrescol, mark=\dresmark,mark size=\teasermarkersize pt, only marks] coordinates {(2.096,31.8)};
    \node [above] at (axis cs:  2.096,31.8) {\textcolor{\asmdrescol}{\tiny R50:$M$\hspace{-0.5pt}$\rightarrow\hspace{-2.5pt}0.35M$}};
%

\addplot[color=\asmnetcol, mark=\netmark,mark size=\teasermarkersize pt, only marks] coordinates {(7.274,36.48)};
\node [left] at (axis cs: 7.274,36.48) {\textcolor{\asmnetcol}{\tiny R50\hspace{-0.5pt}$\rightarrow$\hspace{-.5pt}R18:$M$}};

\end{axis}
\end{tikzpicture}\vspace{-2pt}
  \input{fig/sop_intro}\vspace{-2pt}
  \caption{Retrieval performance (mAP) \vs extraction cost of the query representation (GFLOPs) for Cars196 (top) and SOP (bottom). The notation format used is ``database setup''$\rightarrow$``query setup'', where R50 and R18 are two variants of ResNet architecture. $M$ is equal to 448 and indicates the width and height of images. Contrary to the standard symmetric retrieval (circle), the query in the asymmetric setting is processed by a lighter network (triangle) or in a smaller resolution (diamond, pentagon). \includegraphics[height=1.7ex]{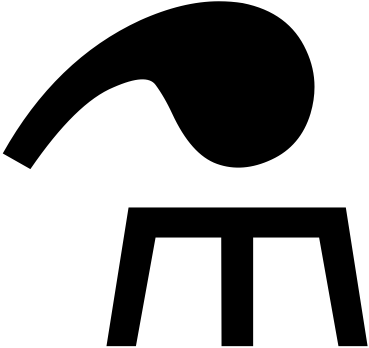}: networks trained with the proposed distillation approach to achieve resolution asymmetry (the focus of this work) and also network asymmetry for comparison. 
  \label{fig:tradeoff}
  \vspace{-12pt}
  }
\end{figure}

\textbf{Performance \vs efficiency.}
We measure the query extraction cost with FLOPs for combinations of networks and specific query resolutions. The trade-off between performance and efficiency for Cars196 and SOP is summarized in Figure~\ref{fig:tradeoff} (in Figure~\ref{fig:cub_intro} for CUB200).
For the case of teacher symmetric testing, the teacher is trained at the resolution it is tested on. In all other cases, the teacher is trained at resolution equal to 448. 
Our distillation approach is applicable in an off-the-shelf way to achieve network asymmetry too. We additionally perform distillation for network asymmetry using only \equ{abs1} (no augmentations) to be as close as possible to AML~\cite{ba+21} for reference.
We observe that resolution asymmetry is outperforming network asymmetry for the same cost and that distillation noticeably outperforms the naive baseline that uses \equ{sarstar}.

\section{Conclusions}
\label{sec:conclusions}
In this work\footnote{Work supported by Junior Star GACR grant No. GM 21-28830M, and Grant Agency of the Czech Technical University in Prague grant No. SGS20/171/OHK3/3T/13.}, we explore resolution asymmetry in deep metric learning and conclude that it forms a better way to optimize the performance \vs efficiency trade-off than network asymmetry studied in prior work. The proposed distillation approach performs well without the use of any labels and allows us to get useful insight into task-tailored augmentations, proper student initialization, and the importance of its different loss terms, namely absolute and relational.
The combination of network and resolution asymmetry is theoretically feasible, possibly even in straightforward ways, but remains as future work. So does the case of dropping the fixed database embeddings assumption and jointly optimizing the database and query networks.

\clearpage
{\small
\bibliographystyle{ieee_fullname}
\bibliography{tex/bib}

\begin{thebibliography}{10}\itemsep=-1pt

\bibitem{asy+19}
Takuya Akiba, Shotaro Sano, Toshihiko Yanase, Takeru Ohta, and Masanori Koyama.
\newblock Optuna: A next-generation hyperparameter optimization framework.
\newblock In {\em {ACM} SIGKDD}, 2019.

\bibitem{bal+16}
Artem Babenko and Victor Lempitsky.
\newblock Efficient indexing of billion-scale datasets of deep descriptors.
\newblock In {\em CVPR}, 2016.

\bibitem{bjw+21}
Yan Bai, Jile Jiao, Shengsen Wu, Yihang Lou, Jun Liu, Xuetao Feng, and Ling-Yu
  Duan.
\newblock Dual-tuning: Joint prototype transfer and structure regularization
  for compatible feature learning.
\newblock In {\em arXiv}, 2021.

\bibitem{bjv+19}
Maxim Berman, Herv{\'e} J{\'e}gou, Vedaldi Andrea, Iasonas Kokkinos, and
  Matthijs Douze.
\newblock {{MultiGrain}: a unified image embedding for classes and instances}.
\newblock In {\em arXiv}, 2019.

\bibitem{bxk+20}
Andrew Brown, Weidi Xie, Vicky Kalogeiton, and Andrew Zisserman.
\newblock Smooth-ap: Smoothing the path towards large-scale image retrieval.
\newblock In {\em ECCV}, 2020.

\bibitem{ba+21}
Mateusz Budnik and Yannis Avrithis.
\newblock Asymmetric metric learning for knowledge transfer.
\newblock In {\em CVPR}, 2021.

\bibitem{chx+19}
Fatih Cakir, Kun He, Xide Xia, Brian Kulis, and Stan Sclaroff.
\newblock Deep metric learning to rank.
\newblock In {\em CVPR}, 2019.

\bibitem{cwq+19}
Ken Chen, Yichao Wu, Haoyu Qin, Ding Liang, Xuebo Liu, and Junjie Yan.
\newblock R3 adversarial network for cross model face recognition.
\newblock In {\em CVPR}, 2019.

\bibitem{ckn+20}
Ting Chen, Simon Kornblith, Mohammad Norouzi, and Geoffrey Hinton.
\newblock A simple framework for contrastive learning of visual
  representations.
\newblock In {\em ICML}, 2020.

\bibitem{dsl+09}
Wei Dong, Richard Socher, Li Li-Jia, Kai Li, and Li Fei-Fei.
\newblock {ImageNet}: A large-scale hierarchical image database.
\newblock In {\em CVPR}, 2009.

\bibitem{dzy+21}
Rahul Duggal, Hao Zhou, Shuo Yang, Yuanjun Xiong, Wei Xia, Zhuowen Tu, and
  Stefano Soatto.
\newblock Compatibility-aware heterogeneous visual search.
\newblock In {\em CVPR}, 2021.

\bibitem{dms+21}
Mihai Dusmanu, Ondrej Miksik, Johannes~L Schönberger, and Marc Pollefeys.
\newblock Cross-descriptor visual localization and mapping.
\newblock In {\em ICCV}, 2021.

\bibitem{evt+20}
Ismail Elezi, Sebastiano Vascon, Alessandro Torcinovich, Marcello Pelillo, and
  Laura Leal-Taix{\'e}.
\newblock The group loss for deep metric learning.
\newblock In {\em ECCV}, 2020.

\bibitem{gzy+19}
Qinquan Gao, Yan Zhao, Gen Li, and Tong Tong.
\newblock Image super-resolution using knowledge distillation.
\newblock In {\em ACCV}, 2019.

\bibitem{gzl+20}
Shiming Ge, Shengwei Zhao, Chenyu Li, Yu Zhang, and Jia Li.
\newblock Efficient low-resolution face recognition via bridge distillation.
\newblock {\em {\sc IEEE} Transactions on Image Processing}, 29:6898--6908,
  2020.

\bibitem{hcl06}
Raia Hadsell, Sumit Chopra, and Yann LeCun.
\newblock Dimensionality reduction by learning an invariant mapping.
\newblock In {\em CVPR}, 2006.

\bibitem{hpt+15}
Song Han, Jeff Pool, John Tran, and William~J. Dally.
\newblock Learning both weights and connections for efficient neural networks.
\newblock In {\em NeurIPS}, 2015.

\bibitem{hzr+16}
Kaiming He, Xiangyu Zhang, Shaoqing Ren, and Jian Sun.
\newblock Deep residual learning for image recognition.
\newblock In {\em CVPR}, 2016.

\bibitem{hdl+20}
Zibin He, Tao Dai, Jian Lu, Yong Jiang, and Shu-Tao Xia.
\newblock Fakd: Feature-affinity based knowledge distillation for efficient
  image super-resolution.
\newblock In {\em ICIP}, 2020.

\bibitem{hvd+15}
Geoffrey Hinton, Oriol Vinyals, and Jeff Dean.
\newblock Distilling the knowledge in a neural network.
\newblock In {\em arXiv}, 2015.

\bibitem{hsc+19}
Andrew Howard, Mark Sandler, Grace Chu, Liang-Chieh Chen, Bo Chen, Mingxing
  Tan, Weijun Wang, Yukun Zhu, Ruoming Pang, Vijay Vasudevan, Quoc~V. Le, and
  Hartwig Adam.
\newblock Searching for mobilenetv3.
\newblock In {\em ICCV}, 2019.

\bibitem{hjl+19}
Jie Hu, Rongrong Ji, Hong Liu, Shengchuan Zhang, Cheng Deng, and Qi Tian.
\newblock Towards visual feature translation.
\newblock In {\em CVPR}, 2019.

\bibitem{hbc+22}
Weihua Hu, Rajas Bansal, Kaidi Cao, Nikhil Rao, Karthik Subbian, and Jure
  Leskovec.
\newblock Learning backward compatible embeddings.
\newblock In {\em arXiv}, 2022.

\bibitem{jds11}
Herv\'e J\'egou, Matthijs Douze, and Cordelia Schmid.
\newblock Product quantization for nearest neighbor search.
\newblock {\em PAMI}, 33(1):117--128, Jan. 2011.

\bibitem{ka14}
Yannis Kalantidis and Yannis Avrithis.
\newblock Locally optimized product quantization for approximate nearest
  neighbor search.
\newblock In {\em CVPR}, 2014.

\bibitem{kaf+20}
Syed~Safwan Khalid, Muhammad Awais, Zhen-Hua Feng, Chi-Ho Chan, Ammarah Farooq,
  Ali Akbari, and Josef Kittler.
\newblock Resolution invariant face recognition using a distillation approach.
\newblock {\em {\sc IEEE} Transactions on Biometrics, Behavior, and Identity
  Science}, 2(4):410--420, 2020.

\bibitem{ksd+13}
Jonathan Krause, Michael Stark, Jia Deng, and Li Fei-Fei.
\newblock 3d object representations for fine-grained categorization.
\newblock In {\em ICCVW}, 2013.

\bibitem{llk+20}
Wonkyung Lee, Junghyup Lee, Dohyung Kim, and Bumsub Ham.
\newblock Learning with privileged information for efficient image
  super-resolution.
\newblock In {\em ECCV}, 2020.

\bibitem{lh+19}
Ilya Loshchilov and Frank Hutter.
\newblock Decoupled weight decay regularization.
\newblock In {\em ICLR}, 2019.

\bibitem{pkl+19}
Wonpyo Park, Dongju Kim, Yan Lu, and Minsu Cho.
\newblock Relational knowledge distillation.
\newblock In {\em CVPR}, 2019.

\bibitem{rtc19}
Filip Radenovi{\'c}, Giorgos Tolias, and Ond{\v{r}}ej Chum.
\newblock Fine-tuning cnn image retrieval with no human annotation.
\newblock {\em PAMI}, 41(7):1655--1668, 2019.

\bibitem{rvp+22}
Vivek Ramanujan, Pavan Kumar~Anasosalu Vasu, Ali Farhadi, Oncel Tuzel, and Hadi
  Pouransari.
\newblock Forward compatible training for large-scale embedding retrieval
  systems.
\newblock In {\em CVPR}, 2022.

\bibitem{rbk+15}
Adriana Romero, Nicolas Ballas, Samira~Ebrahimi Kahou, Antoine Chassang, Carlo
  Gatta, and Yoshua Bengio.
\newblock Fitnets: Hints for thin deep nets.
\newblock In {\em ICLR}, 2015.

\bibitem{rmo+21}
Karsten Roth, Timo Milbich, Bjorn Ommer, Joseph~Paul Cohen, and Marzyeh
  Ghassemi.
\newblock Simultaneous similarity-based self-distillation for deep metric
  learning.
\newblock In {\em ICML}, 2021.

\bibitem{rms+20}
Karsten Roth, Timo Milbich, Samarth Sinha, Prateek Gupta, Bjorn Ommer, and
  Joseph~Paul Cohen.
\newblock Revisiting training strategies and generalization performance in deep
  metric learning.
\newblock In {\em ICML}, 2020.

\bibitem{sxx+20}
Yantao Shen, Yuanjun Xiong, Wei Xia, and Stefano Soatto.
\newblock Towards backward-compatible representation learning.
\newblock In {\em CVPR}, 2020.

\bibitem{st+17}
Leslie~N. Smith and Nicholay Topin.
\newblock Super-convergence: Very fast training of neural networks using large
  learning rates.
\newblock In {\em arXiv}, 2017.

\bibitem{sxj+15}
Hyun~Oh Song, Yu Xiang, Stefanie Jegelka, and Silvio Savarese.
\newblock Deep metric learning via lifted structured feature embedding.
\newblock In {\em arXiv}, 2015.

\bibitem{slj+15}
Christian Szegedy, Wei Liu, Yangqing Jia, Pierre Sermanet, Scott Reed, Dragomir
  Anguelov, Dumitru Erhan, Vincent Vanhoucke, and Andrew Rabinovich.
\newblock Going deeper with convolutions.
\newblock In {\em CVPR}, 2015.

\bibitem{tdt20}
Eu~Wern Teh, Terrance DeVries, and Graham~W Taylor.
\newblock Proxynca++: Revisiting and revitalizing proxy neighborhood component
  analysis.
\newblock In {\em ECCV}, 2020.

\bibitem{tvd+19}
Hugo Touvron, Andrea Vedaldi, Matthijs Douze, and Herve Jegou.
\newblock Fixing the train-test resolution discrepancy.
\newblock In {\em NeurIPS}, 2019.

\bibitem{wcw+22}
Timmy S.~T. Wan, Jun-Cheng Chen, Tzer-Yi Wu, and Chu-Song Chen.
\newblock Continual learning for visual search with backward consistent feature
  embedding.
\newblock In {\em CVPR}, 2022.

\bibitem{wcy+20}
Chi Wang, Ya-Liang Chang, Shang-Ta Yang, Dong Chen, and Shang-Hong Lai.
\newblock Unified representation learning for cross model compatibility.
\newblock In {\em BMVC}, 2020.

\bibitem{wlj+20}
Huan Wang, Suhas Lohit, Michael Jones, and Yun Fu.
\newblock Knowledge distillation thrives on data augmentation.
\newblock In {\em arXiv}, 2020.

\bibitem{wsl+14}
Jiang Wang, Yang Song, Thomas Leung, Chuck Rosenberg, Jingbin Wang, James
  Philbin, Bo Chen, and Ying Wu.
\newblock Learning fine-grained image similarity with deep ranking.
\newblock In {\em CVPR}, 2014.

\bibitem{whh+19}
Xun Wang, Xintong Han, Weilin Huang, Dengke Dong, and Matthew~R Scott.
\newblock Multi-similarity loss with general pair weighting for deep metric
  learning.
\newblock In {\em CVPR}, 2019.

\bibitem{wsl+20}
Yikai Wang, Fuchun Sun, Duo Li, and Anbang Yao.
\newblock Resolution switchable networks for runtime efficient image
  recognition.
\newblock In {\em arXiv}, 2020.

\bibitem{wbm+10}
Peter Welinder, Steve Branson, Takeshi Mita, Catherine Wah, Florian Schroff,
  Serge Belongie, and Pietro Perona.
\newblock Caltech-ucsd birds 200.
\newblock Technical report, California Institute of Technology, 2010.

\bibitem{wms+17}
Chao-Yuan Wu, R Manmatha, Alexander~J Smola, and Philipp Krahenbuhl.
\newblock Sampling matters in deep embedding learning.
\newblock In {\em ICCV}, 2017.

\bibitem{wwz+22}
Hui Wu, Min Wang, Wengang Zhou, Houqiang Li, and Qi Tian.
\newblock Contextual similarity distillation for asymmetric image retrieval.
\newblock In {\em CVPR}, 2022.

\bibitem{yzc+20}
Taojiannan Yang, Sijie Zhu, Chen Chen, Shen Yan, Mi Zhang, and Andrew Willis.
\newblock Mutualnet: Adaptive convnet via mutual learning from network width
  and resolution.
\newblock In {\em ECCV}, 2020.

\bibitem{zgs+22}
Binjie Zhang, Yixiao Ge, Yantao Shen, Shupeng Su, Fanzi Wu, Chun Yuan, Xuyuan
  Xu, Yexin Wang, and Ying Shan.
\newblock Towards universal backward-compatible representation learning.
\newblock In {\em arXiv}, 2022.

\bibitem{zcd+18}
Hongyi Zhang, Moustapha Cisse, Yann~N Dauphin, and David Lopez-Paz.
\newblock mixup: Beyond empirical risk minimization.
\newblock In {\em ICML}, 2018.

\bibitem{zcx+21}
Yiman Zhang, Hanting Chen, Xinghao Chen, Yiping Deng, Chunjing Xu, and Yunhe
  Wang.
\newblock Data-free knowledge distillation for image super-resolution.
\newblock In {\em CVPR}, 2021.

\end{thebibliography}
}
\flushend

\end{document}